\documentclass[a4paper,conference]{IEEEtran}
\ifCLASSINFOpdf
\else
\fi
\usepackage{tikz}
\usetikzlibrary{decorations.pathreplacing}
\usetikzlibrary{fit}
\usetikzlibrary{decorations.markings,bending}
\usepackage{algorithm} 
\usepackage[noend]{algorithmic}
\usepackage{graphicx}
\usepackage{comment}
\usepackage{amsmath,amssymb} 
\usepackage{color}
\usepackage{relsize}

\newcommand\blfootnote[1]{%
  \begingroup
  \renewcommand\thefootnote{}\footnote{#1}%
  \addtocounter{footnote}{-1}%
  \endgroup
}

\begin{document}
%
\title{ClusterFace: Joint Clustering and Classification for Set-Based Face Recognition
}


\author{\IEEEauthorblockN{Samadhi~Wickrama Arachchilage and Ebroul Izquierdo}
\IEEEauthorblockA{Multimedia and Vision Group\\ School of Electronic Engineering and Computer Science\\
Queen Mary University of London, United Kingdom\\
Email: \{s.wickramaarachchilage, ebroul.izquierdo\}@qmul.ac.uk}
}


%


\maketitle

\begin{abstract}
Deep learning technology has enabled successful modeling of complex facial features when high quality images are available. Nonetheless, accurate modeling and recognition of human faces in real world scenarios `on the wild' or under adverse conditions remains an open problem.  When unconstrained faces are mapped into deep features, variations such as illumination, pose, occlusion, etc., can create inconsistencies in the resultant feature space. Hence, deriving conclusions based on direct associations could lead to degraded performance. This rises the requirement for a basic feature space analysis prior to face recognition. This paper devises a joint clustering and classification scheme which learns deep face associations in an easy-to-hard way. Our method is based on hierarchical clustering where the early iterations tend to preserve high reliability.  The rationale of our method is that a reliable clustering result can provide insights on the distribution of the feature space, that can guide the classification that follows. Experimental evaluations on three tasks, face verification, face identification and rank-order search, demonstrates better or competitive performance compared to the state-of-the-art, on all three experiments.
\end{abstract}


%
\IEEEpeerreviewmaketitle

\section{Introduction} 

\blfootnote{© 20XX IEEE.  Personal use of this material is permitted.  Permission from IEEE must be obtained for all other uses, in any current or future media, including reprinting/republishing this material for advertising or promotional purposes, creating new collective works, for resale or redistribution to servers or lists, or reuse of any copyrighted component of this work in other works.}

Face recognition made tremendous advancements over the last decade with a plethora of systems based on deep convolutional neural networks (DCNN). Starting from DeepFace in 2014, face recognition systems consistently reported near-human or even advanced performance on classic datasets like LFW \cite{lfw}. Followed by the performance saturation on LFW, subsequent benchmarks like IJB-A \cite{ijb_a} aimed to address the rather challenging problem of unconstrained face recognition. Consequently, diverse architectural enhancements and learning based solutions addressing issues related to low-quality images, varying pose, illumination changes, emotional expressions, etc., have been proposed and studied over the recent years.
\medskip

Generic advances of face recognition include increasing dataset size, employing more sophisticated deep networks and using ensembles of multiple models etc. Facebook and Google used large in-house datasets of millions in scale to train face recognition models \cite{Facebook, DeepFace}. The success of these systems, resulted in large and openly accessible face datasets like VGGFace2 \cite{vggface2}. Parallel progress was reported in research landscapes such as deeper and sophisticated network architectures and advanced loss functions. These leading contributions act as a strong foundation for novel systems by providing  richer feature representations for faces.
\medskip

Learning based solutions aim to increase the representative power of deep features assuming powerful feature space will be more robust to variations in unconstrained data. However, face recognition under adverse conditions entails specific complications that cannot be easily remedied by generic solutions. These faces are naturally less informative due to blur, low-resolution, poor illumination or insufficient identity cues due to extreme poses. Additionally, these faces could be confusing and even contradictory. For example, the weak samples of different individuals could be of more visual similarity than between a weak sample and a stable sample of the same individual. When such face samples are mapped into deep features, the inconsistencies are carried onto the feature space and hence plain associations based on first-perception could be misleading. While these complications cannot be eliminated entirely, their impact can be minimized by feature distribution learning and neighbourhood analysis. A simple and creative solution can be composed via an easy-to-hard self learning of deep feature distribution where one begins by associating the most confident neighbors and incrementally progress towards confusing samples. 
\medskip

This paper presents a joint clustering and classification scheme code-named `ClusterFace', where clustering is introduced as an intermediary leaning step. During this stage, we gather the closest neighbours into confident clusters. Next we use the cluster base result to classify the less close neighbours. We use this incrementally learnt information,  along with local smoothness assumption to formulate a set of constraints which guide the final recognition task. In doing so, we formulate associations in a gradual and incremental fashion rather than conventional single-prediction based recognition.
\medskip

\begin{figure*}
\centering
\includegraphics[height=7cm]{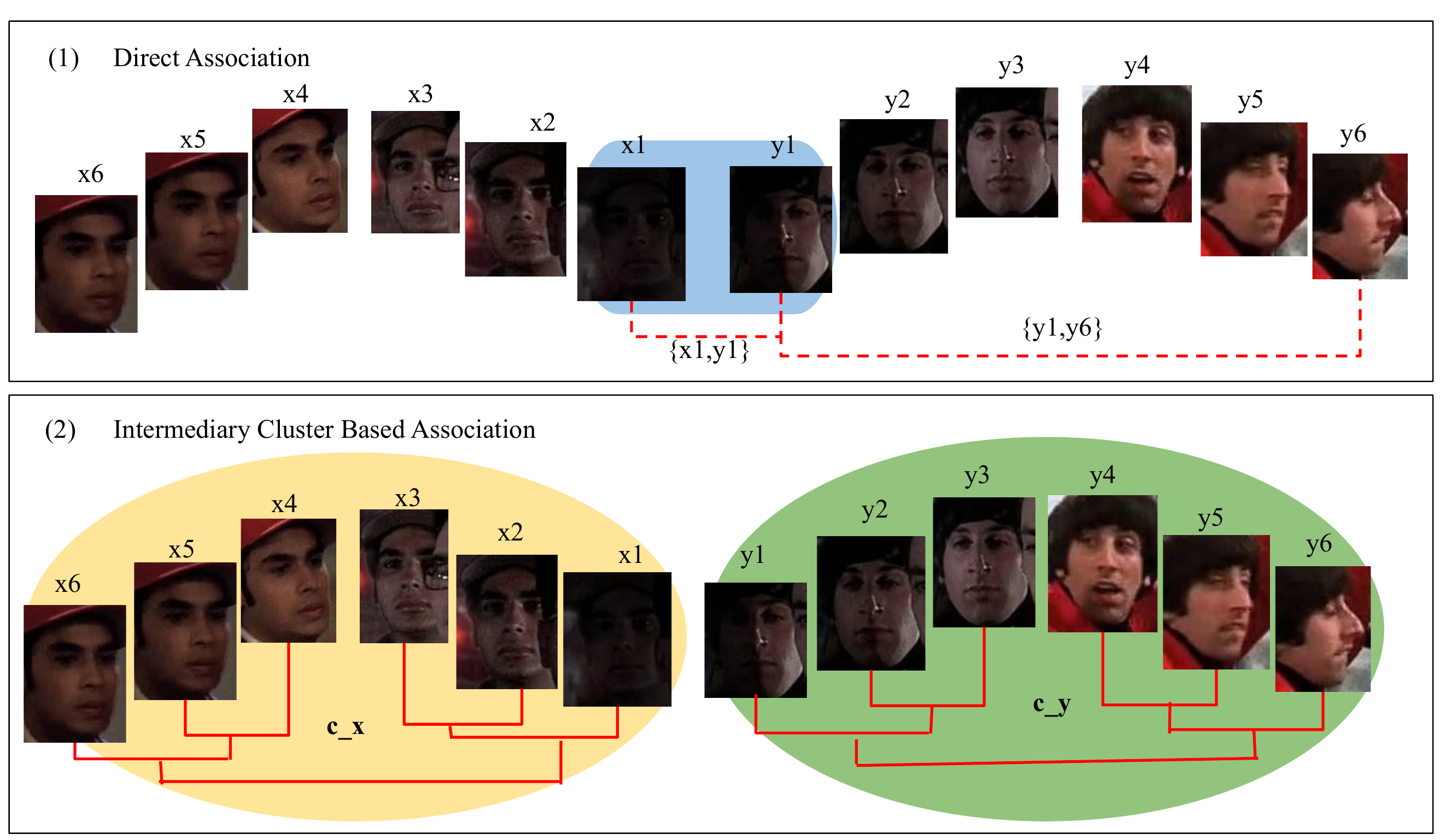}
\caption{The benefit of hierarchical clustering based associations. ${x1,..x6}$ are faces of individual x and ${y1,..., y6}$ are of individual y. $c\_x$ denotes cluster x and $c\_y$ denotes cluster y. Dashed lines show direct associations, solid lines show ClusterFace associations. \textbf{Top}: Direct associations incorrectly conclude that x1 and y1 (faces of different individuals under similar illumination conditions) are more similar than y1 and y6 (faces of same individual under different settings). \textbf{Bottom}: ClusterFace associations begin by merging the closest faces. Gradual and incremental associations correctly group the samples of same individual in to a single cluster.}
\label{fig:clusterface}
\end{figure*}

Figure \ref{fig:clusterface} provides a visual demonstration of ClusterFace association formulation. It exemplifies how misleading information can be restored by incremental learning. This cluster based association is applicable on tasks such as face verification, identification and retrieval. In face verification, ClusterFace assists in identifying faces of the same individual under different environments by chaining through similarities. In face identification and retrieval tasks where a search term probe is compared against a gallery of enrolled faces, we map the probe and the gallery in to the same space and develop ClusterFace learning.  This mapping is particularly helpful in cross-media recognition where  where there is a gap between probe and gallery samples. In such circumstances, ClusterFace  identifies and exploits instances where probe and gallery has clear relationships.
\medskip

The main contributions of the paper are as follows. We present a clustering based deep feature distribution learning scheme which aims to minimize the impact of confusing samples. The solution is composed as a generic-set based recognition scheme which is invariant to the image order, and hence is applicable on both videos and images. On top of the proposed cluster based learning, we formulate face verification, recognition and rank-order search as constrained classification problems. Experimental evaluation on challenging benchmarks show that the proposed approach exploits the deep features in an optimal way.
\medskip

The rest of the paper is organized as follows. Section 2 discusses the related literature on unconstrained face recognition. Section 3 presents the ClusterFace workflow followed by the performance evaluation in section 4. Section 5 presents the conclusion.
\medskip

\section{Related Work} 

The approach we proposed in this paper aims to address unconstrained face recognition by building on advanced deep networks and self formulated constraints. Therefore, we briefly review related works from three aspects, including deep network architectures, unconstrained face recognition and self-formulated constraints.
\medskip

\subsection{DCNN architectures}

Starting  from  LeNet  in  1989 \cite{LeNet5}, DCNNs  have evolved into sophisticated networks particularly fuelled by classification challenges like The ImageNet Large Scale Visual Recognition  Challenge  (ILSVRC) \cite{ImageNet}.  These sophisticated networks were exploited in many computer vision tasks including face recognition. Some  successful face recognition applications that used image classification networks are: DeepId3 \cite{DeepId3} which was influenced by VGGNet \cite{VGGNet}  and GoogLeNet \cite{GoogleNet}, Google's Facenet \cite{FaceNet} which used GoogleNet \cite{GoogleNet}  architecture and VGGFace \cite{VGGFace} that exploited concepts from VGGNet  \cite{VGGNet}. Inspired, we  use Inception ResNet V1 network  \cite{InceptionV4} which has shown to be effective in image recognition with a 4.3\% top-5 error rate in ILSVRC 2012.
\medskip

A  deep  network  is  generally  underpinned by an optimization loss  function which plays an important role in adding  discriminative  power  to the  generated  features.  Over  the  years,  loss  functions have evolved complementing the network architectures. These loss functions can be categorized as classification based approaches, i.e., softmax loss and it's variants \cite{CenterLoss, SphereFace}, and metric learning approaches, i.e.,  contrastive  loss \cite{Contrastive}  and  triplet  loss \cite{FaceNet}. This paper uses a combination of two approaches (i.e., softmax loss for pre-training and triplet loss for further fine-tuning) to achieve high representational power.
\medskip 

\subsection{Unconstrained recognition}

Face recognition under unconstrained settings, requires additional guidance on top of deep features. Popular guidance measures include, using multiple classifiers which guide each other, adaptive self-learning,  guidance through visual or spatio-temporal cues, or strong sample based predictions.
\medskip

Adaptive facial models employ single or multiple classifiers like support vector machines, nearest neighbour classifiers, etc., that are updated based on face tracking results \cite{SSNaive,ssNaive1}. These approaches are applicable only on videos where, the additional spatio-temporal information present in videos can be used to generate face tracks. Co-training is the process of training multiple classifiers where the classifiers guide each other based on confident predictions \cite{coTrainingTrans, SSNaive2}. While co-training has reported considerable performance, their effectiveness mainly depends on the quality and quantity of the labelled data that initiate the classifiers. Bhatt et al. \cite{coTrainingTrans} used an ensemble of  support vector machines trained on labelled gallery and some labelled probe samples. Such semi-supervised approaches require some labelled test data. Co-training and adaptive facial model generation have similarities to ClusterFace in that both approaches  progress through operational data in an easy-to-hard way. However, unlike the other two approaches, ClusterFace initiates by clustering the feature space into unknown number of clusters, and hence requires no additional information or labelled data.
\medskip

 
 
Another approach of minimizing the effect of adverse samples is explicit subset selection, where high quality samples are selected through quality assessment \cite{subset1, subset2, subset3}. While this approach has been proven to be effective through empirical studies, the challenge in this is finding a solid definition for `face quality' \cite{patch}. Yang et al. proposed a neural aggregation network which is trained to automatically advocate high-quality face images while repelling low-quality ones \cite{NAN}. ClusterFace is a much simpler alternative where instead of identifying strong samples, we identify strong relationships. The quality of a relationship is defined in terms of the semantic similarity of corresponding deep features.
 \medskip
  
\subsection{Self formulated constraints}

Self formulated constraints have been used in video based tracking and clustering. These  applications use two pairwise affinity constraints, Cannot-Link (CL) and Must-Link (ML) \cite{simulCT, constrainedClus, Zhang2019}. These constraints are formed exploiting spatio-temporal affinities of multi-face videos where faces within a single shot (a shot is a set of adjacent frames which covers a single motion pattern) are considered must-link and faces in a single frame are considered as cannot link. While these constraints certainly does provide guidance for face tracking and clustering, these are not applicable on tasks such as face recognition on singleton videos (because there are no overlapping faces to form cannot link constraints), face image or image-set based recognition. Unlike the conventional CL and ML constraints, the ClusterFace constraints can be formulated without spatio-temporal relationships. To the best of our knowledge, this is the first face recognition based study which uses self-formulated constraints on un-ordered face image-sets. 
\medskip

\section{ClusterFace Workflow}

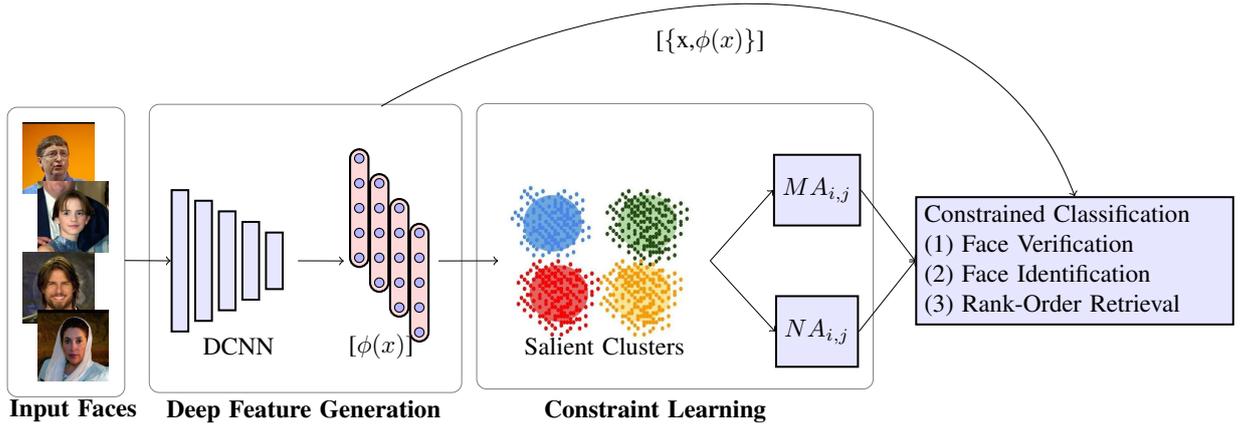
\begin{figure*}[!t]
\centering
\pgfdeclarelayer{background}
\pgfdeclarelayer{foreground}
\pgfsetlayers{background,main,foreground}


\tikzstyle{conv}=[draw,thick,minimum width=0.2cm, fill=blue!10, ]   
\tikzstyle{lossf}=[draw,thick,minimum width=0.2cm, fill=blue!10,  ]  
\tikzstyle{title}=[text width=16em, 
    text centered,]  
\tikzstyle{background}=[rectangle, fill=red!15, inner sep=0.06cm, rounded corners=1.3mm, line width=0.3mm]
\newcommand{\Vcol}[2]{
  \foreach \nn in {1,2,...,5}{
    \node[circle,draw,fill=#2, line width=0.15mm, scale=0.4] (#1-\nn) at (0,{-1*\nn em}) {};
  }
  \begin{pgfonlayer}{background}
    \node [background, fit=(#1-1) (#1-5), draw] (col-#1) {};
  \end{pgfonlayer}
}

\def\blockdist{2.3}
\def\edgedist{2.5}

\resizebox {\linewidth}{!}{
\begin{tikzpicture}

    \node (fc_4) {\includegraphics[height=1 cm]{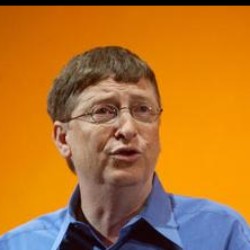}};  
    \path  (fc_4.south)+(0.2,-0.2)  node (bill)   {\includegraphics[height=1 cm]{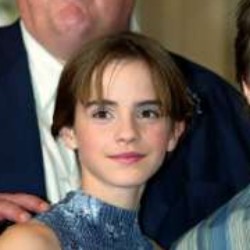}}; 
    \path  (fc_4.south)+(0,-1.2)  node (bill_1)   {\includegraphics[height=1 cm]{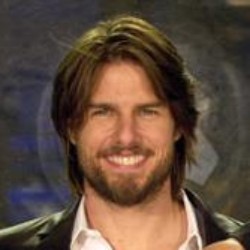}}; 
    \path  (bill.south)+(0,-1.2)  node (bill_2)   {\includegraphics[height=1 cm]{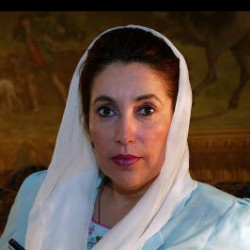}}; 
    
    \path   (bill.south)+(1.5,0) node (c_1) [conv,minimum height=2cm] {};
    \path   (c_1.east)+(0.2,0) node (c_2) [conv, minimum height=1.7cm] {};
    \path   (c_2.east)+(0.2,0) node (c_3) [conv, minimum height=1.4cm] {};
    \path   (c_3.east)+(0.2,0) node (c_4) [conv, minimum height=1.1cm] {};
    \path   (c_4.east)+(0.2,0) node (c_5) [conv, minimum height=0.8cm] {};
    \path   (c_5.east)+(4.5,0) node (clusters)  {\includegraphics[height=2.5 cm]{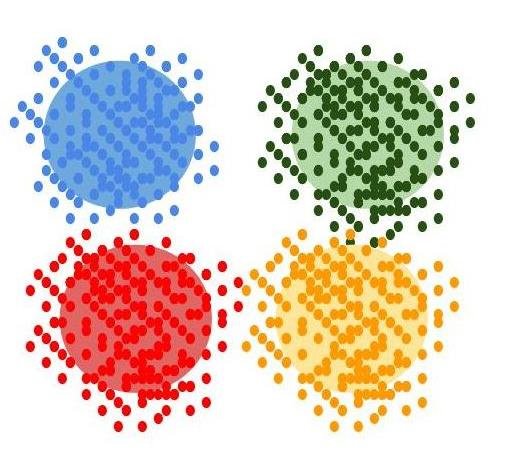}};
    \path   (clusters.east)+(1.5,1) node (cons) [lossf, minimum height=1cm, minimum width=1cm] {$MA_{i,j}$};
    \path   (cons.north)+(-1.5,1.6) node (arrowtxt) {[\{x,$\phi(x)$\}]};
    \path   (clusters.east)+(1.5,-1) node (sa) [lossf, minimum height=1cm] {$NA_{i,j}$};
    \path   (cons.east)+(3,-1) node (loss) [lossf, minimum height=1cm , text width =12em] {Constrained Classification (1) Face Verification \\ (2) Face Identification \\ (3) Rank-Order Retrieval};
    \path (bill_2.south) +(0,-0.3) node (train) [title]  {\textbf{Input Faces}};
    \path (train.east) +(0.3,0) node (train) [title]  {\textbf{Deep Feature Generation}};
    \path (train.east) +(2,0) node (train) [title]  {\textbf{Constraint Learning}};
    \path (c_5.south) +(-0.5,-0.8) node (t_1) [title]  {DCNN};
    \path (c_5.south) +(1.5,-0.8) node (pi) [title]  {[$\phi(x)$]};
    \path (pi.east) +(0.2,0) node (t_1) [title]  {Salient Clusters};
    \draw [->] (pi.north)+(0,3.1) to [out=30,in= 110]  (loss.north)  ;

    \draw[->, black] (bill_1.east)+(0.3,0.38) -- (c_1.west);
    \draw[->, black] (c_5.east)+(0.2,0) -- (4,-1.45);
    \draw[->,  black] (5.33,-1.45) -- (clusters.west);
    \draw[->,  black] (clusters.east) -- (cons.west);
    \draw[->,  black] (clusters.east) -- (sa.west);
    \draw[->,  black] (sa.east) -- (loss.west);
    \draw[->,  black] (cons.east) -- (loss.west);
    
    \begin{scope}[xshift=12em, yshift=1em]
    \Vcol{b}{blue!30}
    \end{scope}
    
    \begin{scope}[xshift=12.8em]
    \Vcol{b}{blue!30}
    \end{scope}
    
    \begin{scope}[xshift=13.6em, yshift=-1em ]
    \Vcol{b}{blue!30}
    \end{scope}
    
    \begin{scope}[xshift=14.4em, yshift=-2em]
    \Vcol{b}{blue!30}
    \end{scope}
    \begin{pgfonlayer}{background}
    \path (clusters.west |- clusters.north)+(-0.3,0.85) node (a) {};
    \path (sa.east |- sa.south)+(0.2,-0.3) node (c) {};
    \path[rounded corners, draw=black!50] (a) rectangle (c);           
    \end{pgfonlayer}

    \begin{pgfonlayer}{background}
    \path (fc_4.west |- fc_4.north)+(-0.1,0.1) node (a) {};
    \path (bill_2.east |- bill_2.south)+(+0.1,-0.1) node (c) {};
    \path[rounded corners, draw=black!50] (a) rectangle (c);           
    \end{pgfonlayer}

    \begin{pgfonlayer}{background}
    \path (c_1.west |- c_1.north)+(-0.3,1.2) node (a) {};
    \path (c_5.east |- c_5.south)+(+2.5,-1.45) node (c) {};
    \path[rounded corners, draw=black!50] (a) rectangle (c);           
    \end{pgfonlayer}

\end{tikzpicture}}
\caption{ The dataflow of the ClusterFace face recognition system. The input faces are the test faces including both gallery and probe images. The pre-trained DCNN model generates feature vectors and the resulting feature space is used to generate a set of salient clusters. The cluster label based constrained matrixes guide the classification that follows.}
\label{System_overview}
\end{figure*}

The problem of face recognition requires association learning among faces. We device a multi-step process which first learns the strong predictions and second builds a set of neighbourhood predictions on top of the strong predictions. These predictions are then used as constraints that guide the remaining predictions. As seen is figure \ref{System_overview}, the process begins with deep feature space generation and is followed by salient clustering, constraint formulation and final recognition.
\medskip

Given a face detection $d$, the deep  network   maps  the  complex  high-dimensional  image information  into a  n-dimensional  proprietary feature  vector $\phi(d) \epsilon R^n$. The  generated  feature  vectors  can  be  interpreted  as  points in a  fixed-dimensional space where the distance between two points is analogous to the level of similarity between the two corresponding faces. The distance measure used in this paper is the cosine similarity between two face detections $d_a,d_b$, which is calculated as follows.
\medskip
 
 \begin{equation}
SIM(d_a,d_b) = \frac{\phi(d_a).{\phi(d_b)}}{||\phi(d_a)||||\phi(d_b)||}
\end{equation}  
\medskip 

Conventional face verification process employs a threshold $\beta$ such that if the distance between two faces is less than $\beta$, they are of the same individual. Building on this convention, we introduce a margin of uncertainty $\gamma$ such that if the  distance between two faces is less than $\beta - \gamma$, they are concluded as of the same individual with higher degree of certainty (i.e strong positive predictions).
\medskip

\subsection{Salient Clustering}

The clustering is based on bottom-up Hierarchical Agglomeration (HAC) which combines the closest cluster in each iteration. Hence, the process exploits two inherent advantages of hierarchical clustering: preserving higher reliability in early iterations and not requiring a pre-specified number of clusters. By setting  $\beta - \gamma$ as the termination height in HAC, this clustering step identifies the most stable associations within the feature space. The clustering algorithm is formulated for generic set-based face recognition tasks where each input face image set $f_i$ is represented by a single point in the deep feature space $\phi(f_i)$. The clustering is detailed in algorithm \ref{GAAC} where the distance between two clusters is measured as the cosine similarity between the two cluster centroids as denoted in the following equation.
\medskip

 \begin{equation}
dist(c_x, c_y) = \frac{(\frac{\sum_{i=1,..N_x}{\phi(x_i)}}{N_x})\text{ . }(\frac{\sum_{i=1,..N_y}{\phi(y_i)}}{N_y} )}{||\frac{\sum_{i=1,..N_x}{\phi(x_i)}}{N_x}||\text{ }||\frac{\sum_{i=1,..N_y}{\phi(y_i)}}{N_y} ||}, 
\end{equation}  
\medskip

where the  two clusters $c_x$ and $c_y$ are of size $N_x$ and $N_y$.  
\medskip

\begin{algorithm}
	\caption{Salient Clustering For Face image sets}
    \hspace*{\algorithmicindent} \textbf{Input:} F: $[\{f_1,\phi(f_1)\},\{f_2,\phi(f_2)\},..., \{f_n,\phi(f_3)]$\}\\
    \hspace*{\algorithmicindent} \textbf{Output:} F where confident positives are merged \\
    \hspace*{\algorithmicindent} \textbf{Initialize:} $dist\_arr = []$ \\ 
    \hspace*{\algorithmicindent}  \hspace*{\algorithmicindent}  \hspace*{\algorithmicindent} \hspace*{\algorithmicindent}   $termination\_dist = \beta-\gamma$\\ 
    
	\begin{algorithmic}[1]
	
	    \STATE {$pairs \leftarrow$ \text{Generate the set of all face-set pairs.}}
	   
    	\FOR {each pair $[f_1,f_2]$ in $pairs$} 
        	\STATE $dist\_arr \leftarrow SIM(f_1,f_2)$
		\ENDFOR
		\STATE Find  $[f\_min_1,f\_min_2]$ corresponding to $min(dist\_arr)$
		
	    \WHILE {$min(dist\_arr) < termination\_dist$}
        	    \STATE $merge\_cluster \leftarrow Merge(f\_min_1, f\_min_2)$  
        	\STATE Delete  $dist\_arr$ elements with   $f\_min_1$ or $f\_min_2$
        	\STATE Delete $f\_min_1, f\_min_2$  from $F$ 
        	\FOR {each face $f$ in $F$}
        	\STATE $dist\_arr \leftarrow dist(f, merge\_cluster)$ 
        	\ENDFOR
        	\STATE $F \leftarrow merge\_cluster$ 
        \ENDWHILE
        \end{algorithmic} 
	\label{GAAC}
\end{algorithm}

\subsection{Constraint Formulation}

We formulate two constraints Must Associate (MA) and Neighbourhood Associate (NA) based on the clustering result, local smoothness assumption and an additional classification step. Local smoothness assumption is imposed in two levels. \textit{The label-level smoothness means that if two observations $x_i$ and $x_j$ are similar, then their labels $y_i$ and $y_j$ should be similar; the constraint-level smoothness tells that given a must-associate between $x_1$ and $x_2$, if $x_3$ is close to $x_2$, then it is assumed that there is also a must-associate between $x_1$ and $x_3$} \cite{constrainedClus}.
\medskip
 
\subsubsection{Must Associate Constraint}

MA constraint states that faces within the same salient cluster are of the same individual. The MA matrix is calculated as follows.
 \begin{equation}
 MA_{i,j} =
\begin{cases}
  c_x, & \text{if}\   i,j \epsilon C_x \\
  NILL, & \text{otherwise}
\end{cases}
\end{equation}  

where $c_x$ is the cluster label of cluster $C_x$.
\medskip

The cluster labels are assigned based on the label-level local smoothness assumption. In particular, all elements within a single salient cluster are identified by a single identity label which is the highest occurring identity. Hence the label of a cluster $C_x$ is given by $MODE\{x_i\}$ where $x_i$ denotes the identity label of cluster element $i$.

\medskip

\subsubsection{Neighbourhood Associate Constraint}

NA is based on the constraint-level local smoothness assumption and k-nearest neighbour search. Given a face $i$, its nearest neighbours $\{n_1, n_2, n_3, ..., n_k\}$  are selected such that $SIM(i,n_1) < SIM(i,n_2), ..., < SIM(i,n_k)$ and subjected to the regularity constraint $SIM(i, n_x) < \beta$. The NA matrix is calculated as follows.

 \begin{equation}
NA_{i,j} =
\begin{cases}
  c_x, & \text{if}\  i \epsilon C_x \text{ AND } N\{j, C_x\} = TRUE\\ 
  NILL, & \text{otherwise}
\end{cases}
\end{equation}  
 
where $c_x$ is the cluster label of cluster $C_x$ and $ N\{j, C_x\}$ is the neighbourhood constraint between a face $j$ and a cluster $C_x$ computed as follows.

\begin{equation}
    \begin{aligned}
 N\{j, C_x\} =  len[(n_1, n_2, ..., n_k) \epsilon C_x] \\ < len[(n_1, n_2, ..., n_k) \epsilon C_y] \\ \text{ for any cluster } C_y \text{ where } y!=x
    \end{aligned} 
\end{equation}
 
where $n_1, n_2, ..., n_k$ are the k-nearest neighbours of $j$ and $[(n_1, n_2, ..., n_k) \epsilon C_x]$ is the sub-set of the k-neighbours that are also elements of cluster $C_x$.

\subsection{Classification}

This section details the three main constrained classification tasks, face identification, verification and rank-order search.
\medskip

\subsubsection{Face verification}

Face verification is the task of determining if two faces belong to the same identity or not. The constrained verification of two faces i and j, can be stated as follows.

  \begin{equation}
V_{i,j} =
\begin{cases}
  True, & \text{if}\    MA_{i,j} \\
  True, & \text{if}\    NA_{i,j} \text{ and } d(m,j)<\beta - \gamma\\
  False, & \text{otherwise}
\end{cases}
\label{veri_con}
\end{equation} 

\subsubsection{Face identification}

Given a probe and a gallery, face identification is the task of matching the probe to its mate in gallery. Given a face $i$, face identification problem is formulated as finding the optimal closest neighbour by a rank order search for rank-1.
\medskip

\subsubsection{Rank-Ordered Results}

Given a probe and a gallery, rand-ordered identification measures what percentage of probe searches return the probe’s gallery mate within the top k rank-ordered results. This is formulated as a constrained ranked search as denoted in algorithm \ref{rank} which ranks the input nearest neighbours ($NN$) of a probe $p$, in the order of similarity closest first.
\medskip

\begin{algorithm}
	\caption{Constrained Rank-Order Search}
    \hspace*{\algorithmicindent} \textbf{Input:} Probe  p\\ 
    \hspace*{\algorithmicindent} \hspace*{\algorithmicindent} \hspace*{\algorithmicindent}
    $MA=\{ MA_{i,j}\}$ \text{  } i,j=\{1,2,...,n\}\\ \hspace*{\algorithmicindent} \hspace*{\algorithmicindent} \hspace*{\algorithmicindent} \hspace*{\algorithmicindent}   $SA=\{ NA_{i,j}\}$\text{  } i,j=\{1,2,...,n\}\\  \hspace*{\algorithmicindent} \hspace*{\algorithmicindent} \hspace*{\algorithmicindent} \hspace*{\algorithmicindent}  $NN=\{ n_1,n_2, ..., n_k\}$\\ 
    \hspace*{\algorithmicindent} \textbf{Output: }  \text{Ranked Neighbours } $RN$\\ 
    \hspace*{\algorithmicindent} \textbf{Initialize: } $RN=[]$
    
	\begin{algorithmic}[1] 
	   
    	\FOR {n in NN} 
        	\IF{$MA_{p,n}$}
        	\STATE $RN \leftarrow n$ \\
        	\STATE DELETE $n$ from $NN$
        	\ENDIF 
		\ENDFOR
        \STATE {$RN \leftarrow x$ $\forall$ $MA_{p,x}$ WHERE $x$ $!\epsilon$ $RN$}
    	\STATE Repeat 1-4 for $\{NA_{i,j}\}$
        \STATE {$RN \leftarrow n$ $\forall$ $n$ $\epsilon$ $NN$}
		 
        \end{algorithmic} 
	\label{rank}
\end{algorithm}

\section {Databases and Evaluations}

The proposed scheme was evaluated in face recognition and verification experiments on two publicly available datasets: including the YouTube and IJB-A database. This section describes the details of the experiments and results.
\medskip

\subsection{Implementation Details}

 We use an implementation of the MTCNN architecture described in \cite{MTCNN} for face detection. As mentioned before, to leverage the recent advent of DCNN architectures, we use Inception ResNet V1 network discussed in \cite{InceptionV4}. The network is trained with VGGFace2 dataset \cite{vggface2} based on softmax loss, and fine-tuned on CASIA-WebFace dataset (overlapping subjects with the evaluation benchmarks removed) \cite{casia} based on triplet loss.
 \medskip 
 
We formulate both video face recognition and set based recognition as a set based recognition problem and perform feature aggregation to assign a single representation to each set of faces. For a more reliable representation, we heavily weigh the stable face appearances using the normalised detection scores, as follows.
 
 \begin{equation}
    \phi(S) = \sum_k w_{s_i}\phi(s_i), \forall s_i \epsilon S 
\end{equation}

where $\phi(S)$ is the set level representation of face image set $S=\{s_1, s_2, ...s_n\}$ of n faces. $w_{s_i}$ is the weight corresponding to the face $s_i$. For image based face recognition, each image is considered as an image-set with cardinality 1.
\medskip

\medskip

\subsection{Face Recognition Benchmarks}
\subsubsection{YouTube Celebrities Face Recognition Dataset}

The YouTube Celebrities (YTC) video dataset consists of 1,910 video sequences of 47 celebrities from YouTube.  There are large variations of pose, illumination, and expression on face videos in this dataset. Moreover, the quality of face videos is very poor because most videos are of high compression rate. The experiment setting is the same as \cite{tyctrans,MMDML}. Five fold cross validation was carried out with three video sequences per subject for training and six for testing in each fold. 
\medskip

\subsubsection{IARPA Janus Benchmark A }
The IJB-A Dataset (IJB-A) \cite{ijb_a} contains 5712 images and 2085 videos of 500 subjects. The average numbers of images and videos per subject are 11.4 images and 4.2 videos. The images are manually aligned as opposed to the general practise of using a commodity  face  detector. The manual alignment process preserves challenging variations such as pose, occlusion, illumination and etc., that are generally filtered out with automated detection. The dataset is a collection of media in the wild which contains both images and videos. 
\medskip

\begin{table}
\begin{center}
\caption{Performance evaluation for verification on IJB-A benchmark. The true accept rates (TAR) vs. false positive rates (FAR).}
\label{ijba}
\begin{tabular}{l|c|c|c}
\hline\noalign{\smallskip}
System & FAR=0.001 & FAR=0.01 & FAR=0.1\\
\noalign{\smallskip}
\hline
\hline
\noalign{\smallskip}

Triplet Similarity \cite{tripSimilarity} & 59.0 & 79.0 & 94.5 \\
Multi-pose (WACV16) \cite{DeepMultiPose} & - & 78.7 & 91.1 \\
Triplet Emb (BTAS16) \cite{tripletEmb} & 81.3 & 91 & 96.4 \\
FastSearch (TPAMI17) \cite{FastSearch} & 51.0& 72.9 & 89.3 \\
Joint Bayesian (WACV16) \cite{jointBaysian} & - & 83.8 & 96.7 \\
PAM (CVPR16) \cite{poseawarecvpr} & 65.2 & 82.6 & - \\
NAN (CVPR16) \cite{NAN} & 88.1  & 94.1 & 97.8  \\
Template (FG17)  \cite{template} & 83.6 &  93.9 & 97.9 \\
DR GAN (CVPR17) \cite{DRGAN}& 53.9 & 77.4 & - \\ 
Contrastive (ECCV18) \cite{contrastiveCNN} & 63.91 & 84.01 & 95.31 \\
\hline 
Direct Associations &  84.95 & 93.76 & 98.13 \\
ClusterFace Associations &  86.60 &94.23 & 98.30 \\ 
\hline
\end{tabular}
\end{center}
\end{table}
\setlength{\tabcolsep}{1.4pt}

\begin{table}
\begin{center}
\caption{Performance evaluation closed-set face recognition on IJB-A benchmark. The percentage identification accuracies in rank-N retrievals}
\label{ijba_id}
\begin{tabular}{l|c|c|c}
\hline\noalign{\smallskip}
System & Rank-1 & Rank-5 & Rank-10\\
\noalign{\smallskip}
\hline
\hline
\noalign{\smallskip}

Triplet Similarity \cite{tripSimilarity} & 88   & 95 &  \\
Deep Multi-pose (WACV 16) \cite{DeepMultiPose} & 84.6 & 92.7 & 94.7 \\
Triplet Emb (BTAS 16) \cite{tripletEmb} & 93.2 & - & 97.7 \\
FastSearch (TPAMI 17) \cite{FastSearch} & 82.2 & 93.1 & - \\
Joint Bayesian (WACV 16) \cite{jointBaysian} & 90.3 & 96.5 &  97.7\\
PAM (CVPR 16) \cite{poseawarecvpr} & 84 & 92.5 & 94.6 \\
NAN (CVPR 16) \cite{NAN} & 95.8  &  98.0  &  98.6  \\
Template (FG 17)  \cite{template} & 92.8 & 97.7  & 98.6 \\
DR GAN (CVPR 17) \cite{DRGAN}& 85.5 & 94.7 & - \\  
\hline

Direct Associations & 94.23 & 97.05 & 97.71\\ 
ClusterFace Associations & 94.28 & 97.05 & 97.72  \\
\hline
\end{tabular}
\end{center}
\end{table}
\setlength{\tabcolsep}{1.4pt}

\setlength{\tabcolsep}{4pt}
\begin{table}
\begin{center}
\caption{Classification Rates (\%) on the
YouTube Celebrities Dataset.}
\label{ytc}
\begin{tabular}{l|c}
\hline
\hline
\noalign{\smallskip}
System & Classification Accuracy (\%)\\
\noalign{\smallskip}
\hline
\noalign{\smallskip} 
SANP (TPAMI 12) \cite{tyctrans}  & 65.60 \\
MMDML (CVPR 15) \cite{MMDML} & 78.5  \\
DRM-PWV  (TPAMI 15) \cite{pami} & 72.55  \\
Fast FR (ICCVW 17) \cite{fastAcc} & 72.1\\
GJRNP (IVC 17) \cite{YANG201747} & 81.3 \\
\hline
Direct Associations  & 90.71 \\
ClusterFace Associations & 91.06 \\ 
\hline
\end{tabular}
\end{center}
\end{table}
\setlength{\tabcolsep}{1.4pt}
\setlength{\tabcolsep}{4pt}

\subsection{Comparison With the Baseline}

We compare the performance levels on ClusterFace associations to the conventional direct associations. This comparison provides a fair evaluation of the CluterFace impact, since both approaches are based on the same deep features. Tables  \ref{ijba} and \ref{ijba_id}, \ref{ytc} reports the evaluation results on  IJB-A and YTC databases. It is clear that the proposed approach achieves improvements over the baseline on all three experiments. In particular, two main observations depicted in the results are (1): the deep features provide a strong foundation for recognition, (2) the Clusterface constraints have guided in correcting a set of   confusing  predictions.
\medskip

\subsection{Comparison With the State-Of-The-Art}

Furthermore, our proposed method is compared with a some of the important state-of-the-art  face recognition systems. Despite all our experiments being carried out on a a simple and plain network without any data augmentation techniques or using multiple ensembles  and under single crop experiments, the proposed approach achieves better or highly competitive results on all occasions. 
\medskip

\subsection{Computational complexity}

The additional clustering step results in an added computational complexity. Since the computational complexity of hierarchical clustering is generally polynomial, the clustering process could be expensive. In particular, given $n$ number of discrete face images, the complexity of hierarchical clustering is generally $O(n^3)$. To avoid such complexity, the presented clustering algorithm replaces each image set with its centroid. Moreover, after each merge, the merged cluster is represented by a single point in the feature space and hence the number of comparisons in the next iteration is reduced. Therefore the computational complexity of the presented algorithm is $O(s^2log(s))$, where s is the number of sets and $s<<n$.  
\medskip

\section{Conclusion}

We have presented a novel face recognition algorithm for matching faces across variations such as pose, illumination and resolution. Associations are learned in an easy-to-hard way where the most confident predictions guide the confusing predictions. The guiding constraints are formulated based on hierarchical clustering. The main advantage of the proposed approach is that there it does not require any additional informational guidance and is solely based on facial features. Hence our approach is easily applicable on a wide range of face recognition based tasks face verification, identification and retrieval.  The usefulness of our algorithms is justified with experiments conducted on the IJB-A database and YTC database in which very good recognition performance is obtained under wide range of pose and illumination and resolution conditions.
\medskip
 
\section*{Acknowledgment}

The research activities leading to this publication has been partly funded by the European Union Horizon 2020 Research and Innovation program under grant agreement No. 786629 (MAGNETO RIA project).






\bibliographystyle{IEEEtran}
\bibliography{ref}
%



\end{document}